# Regularization by Neural Style Transfer for MRI Field-Transfer Reconstruction with Limited Data

**Guoyao Shen[1,2], Yancheng Zhu[1], Mengyu Li[1,2], Ryan McNaughton[1,2], Hernan Jara[3], Sean B. Andersson[1,4], Chad W. Farris[3], Stephan Anderson[2,3], Xin Zhang[1,2,5,6,7*]**

[1]Department of Mechanical Engineering, Boston University, Boston, MA 02215, USA

[2]The Photonics Center, Boston University, Boston, MA 02215, USA

[3]Department of Radiology, Boston University Chobanian & Avedisian School of Medicine, Boston, MA, 02118, USA

[4]Department of Systems Engineering, Boston University, Boston, MA 02215, USA

[5]Department of Electrical & Computer Engineering, Boston, MA, 02215 USA

[6]Department of Biomedical Engineering, Boston, MA, 02215 USA

[7]Division of Materials Science & Engineering, Boston, MA, 02215 USA

*** Correspondence:**
Xin Zhang
xinz@bu.edu

## Abstract

Recent advances in MRI reconstruction have demonstrated remarkable success through deep learning-based models. However, most existing methods rely heavily on large-scale, task-specific datasets, making reconstruction in data-limited settings a critical yet underexplored challenge. While regularization by denoising (RED) leverages denoisers as priors for reconstruction, we propose Regularization by Neural Style Transfer (RNST), a novel framework that integrates a neural style transfer (NST) engine with a denoiser to enable magnetic field-transfer reconstruction. RNST generates high-field-quality images from low-field inputs without requiring paired training data, leveraging style priors to address limited-data settings. Our experiment results demonstrate RNST's ability to reconstruct high-quality images across diverse anatomical planes (axial, coronal, sagittal) and noise levels, achieving superior clarity, contrast, and structural fidelity compared to lower-field references. Crucially, RNST maintains robustness even when style and content images lack exact alignment, broadening its applicability in clinical environments where precise reference matches are unavailable. By combining the strengths of NST and denoising, RNST offers a scalable, data-efficient solution for MRI field-transfer reconstruction, demonstrating significant potential for resource-limited settings.

## 1 Introduction

Magnetic Resonance Imaging (MRI) is a critical medical imaging tool that provides crucial diagnostic insights, significantly influencing clinical decision-making and improving patient outcomes. The continual advancements in MRI technology, particularly the increase in field strength, have led to improved signal-to-noise ratio (SNR), thereby enhancing image quality [Schick et al., 2021; Runge and Heverhagen, 2022]. However, despite its numerous advantages, MRI is inherently limited by long acquisition times. These extended scan durations introduce multiple challenges, including susceptibility to motion artifacts, delays in diagnosis, restricted patient accessibility, and constraints in scanning critically ill patients who may greatly benefit from this imaging technique. To address these limitations, Compressed Sensing (CS) has been introduced as an effective approach, enabling accelerated MRI acquisition by acquiring fewer k-space measurements [Lustig et al., 2007; Donoho, 2006; Candes and Wakin, 2008]. While CS-based methods effectively reduce scan time, they also introduce inherent trade-offs, such as loss of fine image details and potential misalignment artifacts.

To mitigate these issues, recent research efforts have increasingly focused on leveraging deep learning-based techniques for MRI reconstruction [Hammernik et al., 2018; Sriram et al., 2020; Shen et al., 2023; Shen et al., 2024; Hao et al., 2023]. These approaches typically rely on supervised learning with large, high-quality paired datasets to train networks for deblurring and reconstruction tasks [Zbontar et al., 2019; Guo et al., 2024]. However, their dependency on extensive labeled datasets presents a significant limitation, particularly when acquiring high-quality paired data is impractical. This challenge becomes particularly relevant in the context of field-transfer reconstruction, where scans obtained at lower magnetic field strengths need to be reconstructed to mimic high-field MRI images. In situations where high-field scanners are unavailable or scans were originally conducted at lower field strengths, a robust reconstruction technique that can generate high-field quality images from limited data would be highly desirable.

To address inter-scanner and inter-field variations, MRI harmonization techniques have been explored to enhance consistency in quantitative measurements [Stamoulou et al., 2022; Liu et al., 2021]. These methods have shown promising results in reducing disparities between different scanner models and field strengths [Dewey et al., 2020; Wada et al., 2024]. However, many of these approaches still rely on the availability of large-scale paired datasets, making their widespread application challenging. Consequently, the problem of MRI image transformation with limited data remains an open research challenge that requires innovative solutions.

Image domain transfer, a well-studied problem in computer vision, offers a compelling framework for addressing MRI field transfer reconstruction. This technique involves transforming an image to adopt the characteristics of another domain while preserving its core content. Deep learning-based methodologies, particularly Neural Style Transfer (NST) [Gatys et al., 2015a] and Generative Adversarial Networks (GANs) [Goodfellow et al., 2014], have demonstrated remarkable success in image transformation tasks [Chen et al., 2018; Luan et al., 2017; Jing et al., 2020; Azadi et al., 2018]. GAN-based approaches, while effective in generating high-quality images, often suffer from training instability and mode collapse, making them challenging to optimize for limited data applications. Meanwhile, recent advancements in denoising diffusion probabilistic models (DDPMs) [Ho et al., 2020; Saharia et al., 2023] have shown impressive results in generating high-fidelity images with sharper details. However, DDPMs typically require larger datasets to function properly and are more computationally expensive due to their multi-step image generation [Liu et al., 2024]. Moreover, DDPMs poses a significant hindrance in semantically meaningful data representations due to their diffusion process for data deconstruction [Kazerouni et al., 2023]. NST, on the other hand, presents a viable alternative with several unique advantages. Unlike GANs and DDPMs, NST benefits from a stable training process and a flexible network architecture for quick and easy deployment. The core principle of NST involves optimizing two key loss functions: content loss, which preserves the semantic structure of the image, and style loss, which captures the statistical correlations (Gram matrix) of extracted feature maps across multiple network layers. A pre-trained convolutional neural network (CNN), such as VGG, is commonly used as a feature extractor, separating the style transfer process from the feature extraction process. This separation is particularly advantageous in scenarios with limited paired data, as it allows for a more scalable and adaptable pipeline. By leveraging pre-trained feature extractors, NST facilitates effective field-transfer reconstruction without necessitating extensive retraining on domain-specific medical datasets.

Beyond direct image transformation techniques, Regularization by Denoising (RED) has emerged as a powerful framework for image reconstruction by integrating image priors through a denoising engine [Romano et al., 2017]. As an evolution of the Plug-and-Play (PnP) Prior approach [Venkatakrishnan et al., 2013], RED eliminates reliance on Alternating Direction Method of Multipliers (ADMM)-based optimization, offering greater flexibility in selecting denoising algorithms and iterative optimization strategies. This adaptability makes RED particularly suitable for MRI field-transfer reconstruction, as it can effectively handle variations in between low-field and high-field images. Since low-field MRI inherently exhibits increased background noise and altered contrast characteristics due to variations in relaxation times, a reconstruction method that incorporates robust denoising and transformation is crucial for achieving high-quality results. RED has demonstrated impressive performance across various imaging applications, particularly when combined with advanced deep-learning networks [Mataev et al., 2019; Metzler et al., 2018; Wu et al., 2019]. In MRI, RED has been successfully applied to accelerated imaging, motion deblurring, and semi-supervised reconstruction tasks [Gan et al., 2020; Gan et al., 2022; Liu et al., 2020].

Building on these advancements, we propose Regularization by Neural Style Transfer (RNST), an MRI field-transfer reconstruction framework that integrates NST within the RED paradigm, as illustrated in Figure 1. RNST extends RED by incorporating an NST engine with a denoiser, leveraging the pre-trained CNN feature extractor for style transfer without requiring extensive paired training data. This enables RNST to perform effective field-transfer reconstruction with limited data availability, making it highly suitable for real-world clinical applications where paired high-field and low-field datasets are scarce.

As the NST engine uses a pre-trained CNN-based feature extraction network with general image contents, RNST provides a plug-and-play solution that can be adapted to different MRI settings without requiring domain-specific retraining. We validated the effectiveness of RNST using multiple MRI datasets obtained from different scanning conditions, evaluating its ability to reconstruct high-field quality images from limited data while mitigating noise-induced artifacts. Our results demonstrate that RNST achieves superior reconstruction quality with limited data availability. By leveraging the complementary strengths of RED and NST, RNST provides a novel, flexible, and efficient solution for MRI field-transfer reconstruction, addressing the challenge of transforming MRI images across different field strengths without the need for large-scale paired datasets.

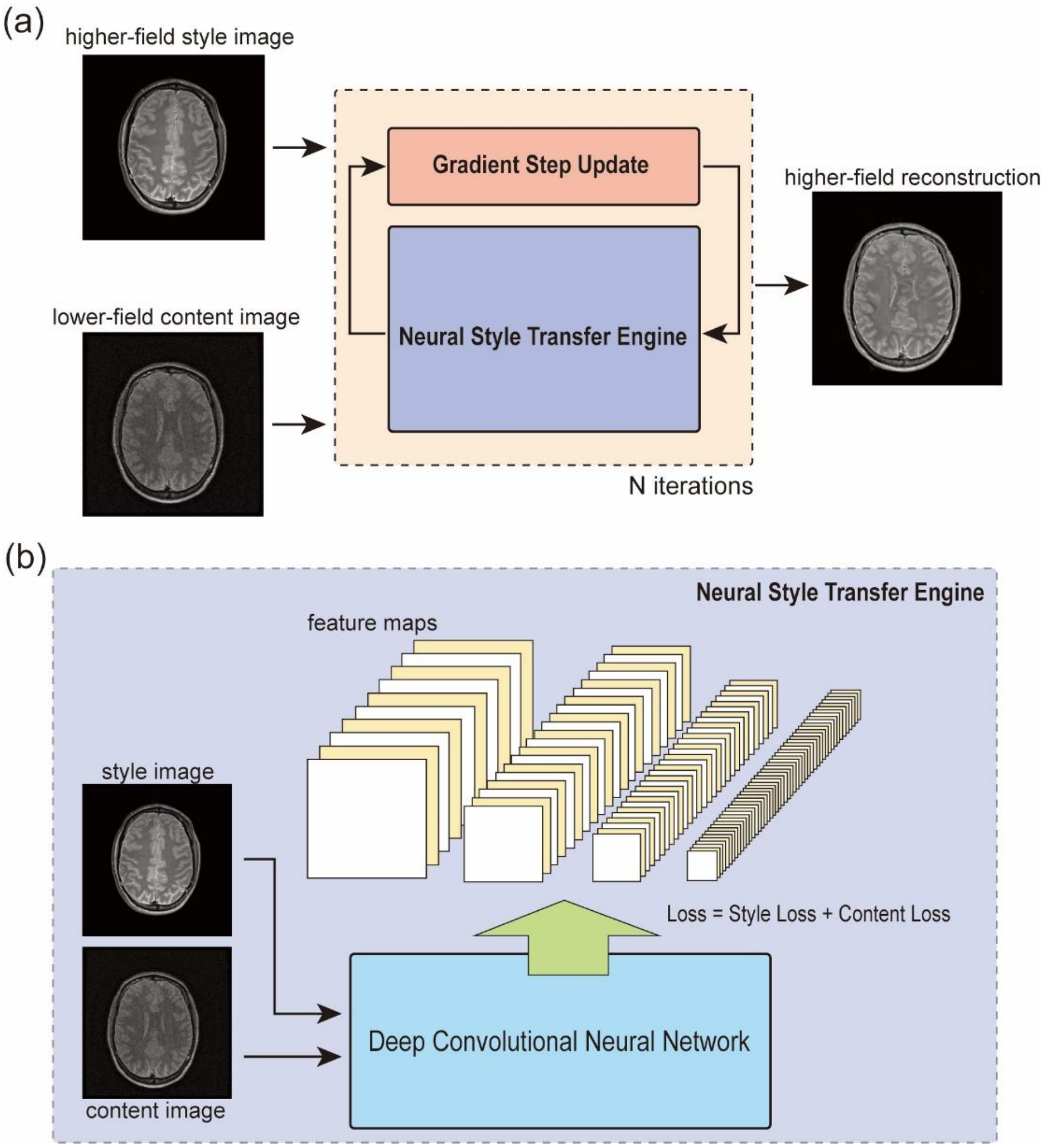

**Figure 1.** The overall framework of regularization by neural style transfer (RNST). **(a)** shows the overall structure of RNST. It contains two main parts in the optimization iteration. The first one is a neural style transfer (NST) network that provides a set of style transferred images from the input. Then, a gradient step update is applied for denoising and reconstruction. The final reconstruction is generated after N iterations. **(b)** demonstrates a closer look at the NST engine. It takes a style image as guidance and a content image for reconstruction. The loss is a combination of the style loss measuring the feature correlations among multiple layers and a content loss measuring the content difference between the output and content feature maps.

## 2 Materials and Methods

### 2.1 Neural style transfer (NST)

NST is a paradigm of deep learning-based style and content separation and recombination. Consider a content image $x_c$ and a style image $x_s$, NST seeks to give an output image $x_{comb}$ which is the combination of the content and style images [Gatys et al., 2015a]. A deep convolutional neural network typically consists of layers of computational units which process visual information hierarchically. The output from a certain layer includes a branch of feature maps. This hierarchically organized network provides a computational representation of the input image where lower layers capture pixel value details and textures while higher layers capture general image contents and shapes [Zhang and Zhu, 2018; Mahendran and Vedaldi, 2015; Gatys et al., 2015b].

NST utilizes a CNN network $U_{CNN}$ to separate the style and content of the original images and recombine them in the output image so that $x_{comb}$ is close to the $x_c$ content-wise, while close to $x_s$ style-wise. More specifically, consider the feature maps of an image $x$ in layer $l$ where they consist of $N_l$ maps in total and each map has the size $M_l$. In this case, all feature maps can be represented by a matrix $F^l \in \mathcal{R}^{N_l \times M_l}$ where $F_{ij}^l$ corresponds to the $i$ th feature map at position $j$. The content loss between the content image $x_c$ and input image $x$ in layer $l$ is defined as the squared-error loss between their feature representations:

$$\mathcal{L}_{content}(x, x_c, l) = \frac{1}{2} \sum_{i,j} \left( F_{ij}^l - F_{c\,ij}^{\,l} \right)^2 \tag{1}$$

Style loss represents the correlations between different feature maps. The correlation is given by the Gram matrix $G^l \in \mathcal{R}^{N_l \times N_l}$, where $G_{ij}^l$ is the inner product of two feature maps $i$ and $j$ in layer $l$:

$$G_{ij}^l = \sum_k F_{ik}^l F_{jk}^l \tag{2}$$

Thus, the matching of the style for a given image in a certain layer is done by minimizing the mean-squared loss between the entries of Gram matrices from the style image and input image:

$$\mathcal{L}_{style}(x, x_s, l) = \frac{1}{4 N_l^2 M_l^2} \sum_{i,j} \left( G_{ij}^l - G_{s\,ij}^{\,l} \right)^2 \tag{3}$$

Then, the style loss among multiple layers is:

$$\mathcal{L}_{style}(x, x_s) = \sum_{l} \omega_l \, \mathcal{L}_{style}(x, x_s, l) \tag{4}$$

where $\omega_l$ is the weighting factor representing the contribution of each layer. The total loss is the combination of content loss and style loss:

$$\mathcal{L}_{total}(x, x_c, x_s) = \alpha\mathcal{L}_{content}(x, x_c) + \beta\mathcal{L}_{style}(x, x_s) \tag{5}$$

where $\alpha$ and $\beta$ are the weighting factors of content and style loss, respectively.

### 2.2 Regularization by denoising (RED)

RED [Romano et al., 2017] provides a flexible pipeline for image reconstruction. Consider a classic reconstruction case where:

$$\boldsymbol{y} = \boldsymbol{H}\boldsymbol{x} + \boldsymbol{e} \tag{6}$$

here $\boldsymbol{H}$ is a degradation operator and $\boldsymbol{e}$ is the additional noise. $\boldsymbol{x}$ represents the unknown reconstruction target and $\boldsymbol{y}$ is the noisy measurement. A typical reconstruction brings the following form:

$$\widehat{\boldsymbol{x}} = \mathrm{argmin}_x \, \ell(\boldsymbol{y}, \boldsymbol{x}) + \lambda\rho(\boldsymbol{x}) \tag{7}$$

where $\widehat{\boldsymbol{x}}$ is the estimated reconstruction of $\boldsymbol{x}$, and $l$ and $\rho$ are penalty and regularization terms. This form includes a branch of image reconstruction tasks such as denoising, deblurring, super-resolution, tomographic reconstruction, and so on. The noise contamination of the measurements can also be probability distributions such as Gaussian, Laplacian, or Poisson depending on the setting.

Previous work such as the PnP prior [Venkatakrishnan et al., 2013] algorithm gives the reconstruction in a block-coordinate-descent fashion where one step is for solving the inverse problem and the other step is for denoising the updated reconstruction. While PnP prior does not specifically refer to a certain choice of the denoising engine as a prior, it comes with the limitation of activating a denoising algorithm and departing from the original setting without an underlying cost function. As the name suggests, regularization by denoising advocates the regularization term as:

$$\rho(\boldsymbol{x}) = \frac{1}{2}\boldsymbol{x}^T\big(\boldsymbol{x} - f(\boldsymbol{x})\big) \tag{8}$$

where $f$ refers to the denoising engine. In this way, RED comes with much more flexibility for the choice of the optimization method and denoising engine.

### 2.3 Regularization by neural style transfer (RNST)

In this section, we demonstrate the RNST method for magnetic field transfer reconstruction. RNST includes a neural style transfer and a denoising engine. The magnetic field transfer reconstruction from a lower-field image to a higher-field requires a process of denoising without loss of features in the tissues. However, since the original image was obtained with a lower magnetic strength, the image quality and noise level are much worse compared to the higher-field one. Though the denoising of background noise can be achieved by a denoising engine, the shifting in contrast ratio and feature loss in the reconstruction still exists. Thus, we employ an NST engine as part of our regularization optimizer to update the lower-field images iteratively such that the correlations between different features become as close as possible to the higher-field references.

Consider a magnetic field transfer reconstruction with the form of (6):

$$\boldsymbol{x} = \boldsymbol{H}\boldsymbol{x}_{\boldsymbol{h}} + \boldsymbol{e} \tag{9}$$

where $\boldsymbol{x}_{\boldsymbol{h}}$ is the unforeseen higher-field target and $\boldsymbol{x}$ is the lower-field noisy measurement. As shown in equation (7) and (8), this reconstruction process can be written in the form:

$$\widehat{\boldsymbol{x}} = \mathrm{argmin}_{x}\ \ell(\boldsymbol{x}_{\boldsymbol{h}}, \boldsymbol{x}) + \tau \boldsymbol{x}^{T}\big(\boldsymbol{x} - D(\boldsymbol{x})\big) \tag{10}$$

where the latter term is a regularization term with a denoiser $D$ integrated and $\tau$ is the corresponding weighting factor. Notice that although the degradation operator $\boldsymbol{H}$ can be hard to define since modeling the imaging process from different magnetic strengths and setups is difficult, the higher-field target $\boldsymbol{x}_{\boldsymbol{h}}$ can be implicitly represented by a guidance scan $\boldsymbol{x}_{\boldsymbol{guid}}$ coming from the same magnetic field. Thus, (10) can be solved by embedding an NST engine in the reconstruction pipeline.

The overall structure of our RNST is demonstrated in Figure 1(a). As an optimizer to reconstruct the low-field input image, it contains two main parts in the optimization iteration. The first one is an NST network that provides a set of directional style transferred images from the raw input. As mentioned above, the NST engine works based on the handling of the content image and style image as shown in Figure 1(b). By computing the style and content loss between the input and style guidance image, the NST engine updates the input with respect to this loss combination. After a number of iterations, the output contains content of the original input but has a feature style, or pixel correlations closer to the style image. NST benefits from the fact that it works based on a deep convolutional neural network usually pre-trained on a large-scale dataset such as ImageNet and the network is frozen for feature extraction during the style transfer process. However, the image contents in our work are different from these pre-trained datasets and might lead to a mismatch in feature extraction. Considering this, we applied an online update with the NST engine to search for directions of our gradient descent optimizer. This online update generates multiple candidates from the NST engine with different style transfer levels and these output images with different style transfer levels play the role of guidance for the gradient evaluation. The second part is a line-search gradient descent engine [Stanimirović and Miladinović, 2010] as an iterative approach for reconstruction. Newton's method provides a faster convergence speed than the classic gradient descent method, yet it requires the calculation of a higher order derivative of the objective function [Deuflhard, 2005]. However, since our reconstruction optimization contains the NST engine outputs and it can be hard to define a numeric derivative of the objective function, we employ a line-search gradient descent as an approximation.

The pseudo-code of our RNST via the line-search gradient descent is formulated as Algorithm 1. Beginning with a noisy lower-field raw input $\boldsymbol{x_{in}}$ and a higher-field style guidance $\boldsymbol{x_{guid}}$, the denoiser $D$ first generates a denoised image $\boldsymbol{x_d}$ from the input. Then, a list of style transferred images $\boldsymbol{x_t}$ are given by the NST engine $\mathcal{T}$. Here the subline index $N_0$ is an initial number and $N_{step}$ is the step size increase for the iteration number of $\mathcal{T}$. After preparation of the style transferred image list $\boldsymbol{L_t}$ , a line-search gradient descent is implemented to find the best gradient descent direction with respect to the objective loss $\mathcal{L}$. In order to overcome the potential convergence to a local optima of the non-convex objective function, we scan the possible solutions based on the list of style transferred images $\boldsymbol{x_t}$, and apply a line-search of different step-sizes $\tilde{\mu}$ as a further exploration. With a batch of candidates in the list covering multiple step sizes, the best one is selected from the list with respect to the objective function in each iteration. Note that the step-size $\tilde{\mu}$ can be adjusted dynamically per iteration. For instance, by applying an Armijo step-size rule [Armijo, 1966], the value of $\tilde{\mu}$ can be updated with respect to the estimation of local gradient and the objective function. Herein, to keep things simple, we set $\tilde{\mu} = i\mu$. The gradient direction is calculated by combining the gradient of the style transfer image $\boldsymbol{x_t}$ and the denoising image $\boldsymbol{x_d}$ to produce an intermediate candidate $\tilde{\boldsymbol{x}}$:

$$\tilde{\boldsymbol{x}} = \boldsymbol{x} - \tilde{\mu}\big(\boldsymbol{x} - \boldsymbol{x_t} + \lambda(\boldsymbol{x} - \boldsymbol{x_d})\big) \tag{11}$$

To evaluate the performance of candidate $\tilde{\boldsymbol{x}}$, a one-step neural style transfer loss is calculated:

$$\begin{aligned} \tilde{\boldsymbol{x}}' &= \mathcal{T}_{N_0'=1}(\tilde{\boldsymbol{x}}, \boldsymbol{x_{guid}}) \\ \mathcal{L}(\tilde{\boldsymbol{x}}') &= \alpha\mathcal{L}_{content}(\tilde{\boldsymbol{x}}', \boldsymbol{x_{in}}) + \beta\mathcal{L}_{style}\big(\tilde{\boldsymbol{x}}', \boldsymbol{x_{guid}}\big) \end{aligned} \tag{12}$$

And the best candidate is kept as the input for the next iteration.

**Algorithm 1:** RNST via Line-search Gradient Descent

1: **Input:** Given $\lambda$, $\mu$, $\alpha$, $\beta$, $\boldsymbol{x_{in}}$, $\boldsymbol{x_{guid}}$
2: **Initialization:** $\mathcal{D}$, $\mathcal{T}$
3: $\boldsymbol{x} \leftarrow \boldsymbol{x_{in}}$
4: **for** $k = 1,2,\ldots, N_{iter}$ **do:**
5: $\boldsymbol{x_d} = \mathcal{D}(\boldsymbol{x})$
6: $\mathcal{L}^* = \mathcal{L}_0$
7: $\boldsymbol{L_t} = \{\}$
8: **for** $j = 1,2,\ldots, N_{style}$ **do:**
9: $\boldsymbol{x_t} = \mathcal{T}_{N_0+jN_{step}}(\boldsymbol{x_d}, \boldsymbol{x_{guid}})$
10: $\boldsymbol{L_t}$ add $\boldsymbol{x_t}$
11: **end for**
12: Line-search Gradient Descent:
13: **for** $i = 1,2,\ldots, N_{line}$ **do:**
14: $\tilde{\mu} = i \cdot \mu$
15: **for** $x_t$ in $L_t$ **do:**
16: $\tilde{\boldsymbol{x}} = \boldsymbol{x} - \tilde{\mu}\big(\boldsymbol{x} - \boldsymbol{x_t} + \lambda(\boldsymbol{x} - \boldsymbol{x_d})\big)$
17: $\tilde{\boldsymbol{x}}' = \mathcal{T}_{N_0'=1}(\tilde{\boldsymbol{x}}, \boldsymbol{x_{guid}})$
18: $\mathcal{L}(\tilde{\boldsymbol{x}}') = \alpha\mathcal{L}_{content}(\tilde{\boldsymbol{x}}', \boldsymbol{x_{in}}) + \beta\mathcal{L}_{style}\big(\tilde{\boldsymbol{x}}', \boldsymbol{x_{guid}}\big)$

19: **if** $\mathcal{L}(\widetilde{\boldsymbol{x}}') < \mathcal{L}^*$ **then**
20: $\boldsymbol{x_{out}} = \widetilde{\boldsymbol{x}}$
21: $\mathcal{L}^* = \mathcal{L}(\widetilde{\boldsymbol{x}}')$
22: **end if**
23: **end for**
24: **end for**
25: $\boldsymbol{x} = \boldsymbol{x_{out}}$
26: **end for**
27: **Result:** The output is $\boldsymbol{x}$

**Algorithm 1.** Pseudo-code of our RNST via line-search gradient descent algorithm.

## 2.4 Model comparisons and implementation details

RNST offers great flexibility in selecting the underlying deep neural network structure for style transfer. In this work, we explored three different network architectures as our NST engine for performance comparison: VGG16 [Simonyan and Zisserman, 2015], ResNet50, and ResNet152 [37 He et al., 2016]. These models, pre-trained on large-scale general visual recognition tasks [Gatys et al., 2015a], are readily accessible through PyTorch [Paszke et al., 2019].

For feature extraction in VGG16, we utilized the first eight layers for computing the style loss and the fourth layer for the content loss. For ResNet50 and ResNet152, style loss calculations were based on features extracted from the 7×7 convolutional layer and the 3×3 convolutional bottleneck layers 1, 2, and 3, while content loss was computed using features from the 7×7 convolutional layer and the 3×3 convolutional bottleneck layers 1 and 3. Additionally, we observed that substituting the traditional L2 loss with L1 loss in both content and style loss computations enhanced the sharpness of the reconstructed images, leading to improved performance.

We incorporated the widely used Block-Matching and 3D Filtering (BM3D) algorithm [Dabov et al., 2007] as the denoising engine. For the RNST algorithm, we set $N_{style} = 3$ and $N_{line} = 5$. The NST engine directional steps were set to $N_0 = 500$ and $N_{step} = 100$. The weighting factor ratio in the NST engine was set to $\alpha//\beta = 10^{-6}$ for VGG16 and $\alpha//\beta = 10^{-4}$ for ResNet50 and ResNet152.

## 2.5 Dataset details and evaluation metrics

We evaluated our method using two datasets: one from the National Alzheimer's Coordinating Center (NACC) and another from our institution. Institutional Review Board (IRB) approval was obtained for this study and informed consent was obtained from the participant imaged in this study. The NACC dataset includes 22 scans from 5 ADRCs (Alzheimer's Disease Research Center) conducted between January 2000 and January 2019, acquired on 1.5T and 3T scanners, with six 3D scans reconstructed in axial, coronal, and sagittal planes. The remaining scans were reconstructed in the axial plane based on their original acquisition. The institutional dataset comprised scans obtained on both 3T and 1.5T scanners (Ingenia Philips Healthcare). The measurements were taken using the ELGAN-ECHO MRI protocol [McNaughton et al., 2022] for the same subject in both magnetic strengths. It included two concatenated scans with identical geometry and receiver settings implemented, which is called a dual-echo turbo spin-echo (TSE) and a single-echo TSE, combined as a triple TSE. The scanning is a triple-weighting acquisition including directly acquired (DA) image 1 for proton density-weighted, DA2 for T2-weighted and DA3 for T1-weighted, voxel = $0.5 \times 0.5 \times 2$mm. Echo times = 12 msec, 102 msec for the first and second effective echo; long repetition time = 10 seconds, short repetition time = 5 seconds. Each DA generated 80 slices, leading to 240 slices for each magnetic strength.

We used the unregistered 3T scan as the style guidance and the 1.5T scan as the content images. The resolution for each slice is 256×256 for the NACC dataset and 512×512 for the triple TSE dataset. We then performed a 3D registration on the 3T scan corresponding to the 1.5T scan to give the registered 3T scan using 3DSlicer [Pieper et al., 2004; Fedorov et al., 2012]. This registered 3T scan worked as the reconstruction reference in our performance evaluation. For the NACC dataset, reconstruction tasks were performed with images further corrupted by additive white Gaussian noise (AWGN) at a level of 0.08 ($\sigma = 20/255$). The number of iterations $N_{iter}$ was set to 10 for ResNet50 and ResNet152, and 30 for VGG16. For all models, the parameters were set to $\mu = 0.1$ and $\lambda = 0.3$.

For the triple TSE dataset, two reconstruction tasks were performed. The first utilized the original 1.5T and 3T scans, while the second incorporated additive white Gaussian noise (AWGN) at a level of 0.08, consistent with the NACC dataset. For VGG16, $N_{iter}$ was set to 10 with $\mu = 0.13$ and $\lambda = 0.2$ for the first task, and to 50 with $\mu = 0.15$ and $\lambda = 0.3$ for the second. For ResNet50 and ResNet152, $N_{iter}$ was set to 30 with $\mu = 0.12$ and $\lambda = 0.2$for both tasks. Our RNST magnetic field transfer reconstruction includes a matched guidance and frozen guidance setup. In the matched guidance setup, the slice number of the guidance image and noisy content image were matched. Note that their image contents were still quite different due to the subject movement. To further demonstrate that RNST benefits from the fact that the guidance $\boldsymbol{x_{guid}}$ encodes the image style and implicitly represents the reconstruction, we froze the guidance image index to $i_{guid} = 55$ and performed reconstruction on the truncated brain portion of slices $i_{brain} = [40,60]$. During the evaluation, each reconstruction $\hat{\boldsymbol{x}}$ was compared to the registered 3T scan with the matched slice index $\boldsymbol{x_{registered}}$. Our quantitative metrics include peak signal-to-noise ratio (PSNR) in dB and structural similarity (SSIM):

$$PSNR(\hat{x}, x) = 10log_{10}\frac{max(x)^2}{MSE(\hat{x}, x)} \tag{13}$$

where $MSE(\hat{x}, x)$ is the mean squared error between the reconstructed image $\hat{x}$ and the reference image $x$.

$$SSIM(\hat{x}, x) = \frac{(2\mu_{\hat{x}}\mu_x + c_1)(2\sigma_{\hat{x}x} + c_2)}{\left(\mu_{\hat{x}}^2 + \mu_x^2 + c_1\right)\left(\sigma_{\hat{x}}^2 + \sigma_x^2 + c_2\right)} \tag{14}$$

where $\mu_{\hat{x}}$, $\mu_x$, $\sigma_{\hat{x}}^2$, and $\sigma_x^2$ are the mean and variance of reconstructed image $\hat{x}$ and reference image $x$, respectively. $\sigma_{\hat{x}x}$ is the covariance of $\hat{x}$ and $x$. $c_1 = (k_1 L)^2$, $c_2 =$ $(k_2L)^2$ are two factors to stabilize the division. $L$ is the dynamic range of pixel-values. We use a window size of $7 \times 7$ with $k_1 = 0.01$ and $k_2 = 0.03$.

## 3 Results

### 3.1 MRI field-transfer reconstruction performance

Table 1 summarizes the performance metrics for the NACC dataset, where all networks achieved remarkable performance for MRI field-transfer reconstruction. Overall, ResNet50 and ResNet152 provide better performance compared to VGG16. For instance, in the coronal plane, ResNet152 achieved a PSNR/SSIM of 22.00/0.7406, compared to ResNet50 at 21.95/0.7165 and VGG16 at 20.26/0.6273. Similarly, in axial plane reconstructions across all scans, ResNet152 achieved a PSNR/SSIM of 21.12/0.6461, compared to ResNet50 at 21.15/0.6271 and VGG16 at 20.50/0.5800. Figure 2 provides qualitative comparisons of reconstructed images across axial, coronal, and sagittal planes. The reconstructed images are noticeably cleaner and more similar in contrast and intensity to those of the reference images when compared to the input lower-field MRI scans.

Similarly, when being applied to the triple TSE dataset, RNST also demonstrated superior performance. Table 2 reports the reconstruction metrics on the triple TSE dataset, evaluated on original scans as well as scans with additional noise. Reconstruction on DA2 with ResNet152 achieved a PSNR/SSIM of 24.30/0.7875. When further corrupted with noises, RNST with ResNet152 still maintained a stable performance at 24.27/0.7820. These results highlight the flexibility of our RNST framework which can be applied with various deep neural networks, achieving stable performance under multiple scanning setups.

**Table 1.** Evaluation metrics of RNST reconstructions on the NACC dataset.

| | Axial | | Coronal | | Sagittal | | All scans (Axial) | |
|---|---|---|---|---|---|---|---|---|
| | PSNR | SSIM | PSNR | SSIM | PSNR | SSIM | PSNR | SSIM |
| Input | 20.57 | 0.2162 | 19.79 | 0.2044 | 19.40 | 0.2572 | 18.55 | 0.1826 |
| VGG16 | 24.06 | 0.7074 | 20.26 | 0.6273 | 20.96 | 0.5951 | 20.50 | 0.5800 |
| ResNet50 | 24.38 | 0.7687 | 21.95 | 0.7165 | 21.37 | 0.6426 | 21.15 | 0.6271 |
| ResNet152 | 24.37 | 0.7882 | 22.00 | 0.7406 | 21.40 | 0.6412 | 21.12 | 0.6461 |

**Table 2.** Evaluation metrics of RNST reconstructions on the triple TSE scans.

| | PSNR | | | SSIM | | |
|---|---|---|---|---|---|---|
| | DA1 | DA2 | DA3 | DA1 | DA2 | DA3 |
| | All slices, original scan | | | | | |
| Input | 20.48 | 20.34 | 20.81 | 0.2659 | 0.2327 | 0.2427 |
| VGG16 | 21.04 | 23.61 | 22.99 | 0.7605 | 0.7716 | 0.7748 |
| ResNet50 | 21.60 | 24.23 | 23.89 | 0.7782 | 0.7813 | 0.7833 |
| ResNet152 | 21.66 | 24.30 | 23.97 | 0.7843 | 0.7875 | 0.7859 |
| | All slices, additional noise | | | | | |
| Input | 18.39 | 18.29 | 18.53 | 0.1055 | 0.1026 | 0.0882 |
| VGG16 | 20.75 | 22.81 | 22.37 | 0.7608 | 0.7664 | 0.7698 |
| ResNet50 | 21.58 | 24.20 | 23.90 | 0.7705 | 0.7727 | 0.7804 |
| ResNet152 | 21.62 | 24.27 | 23.96 | 0.7778 | 0.7820 | 0.7854 |

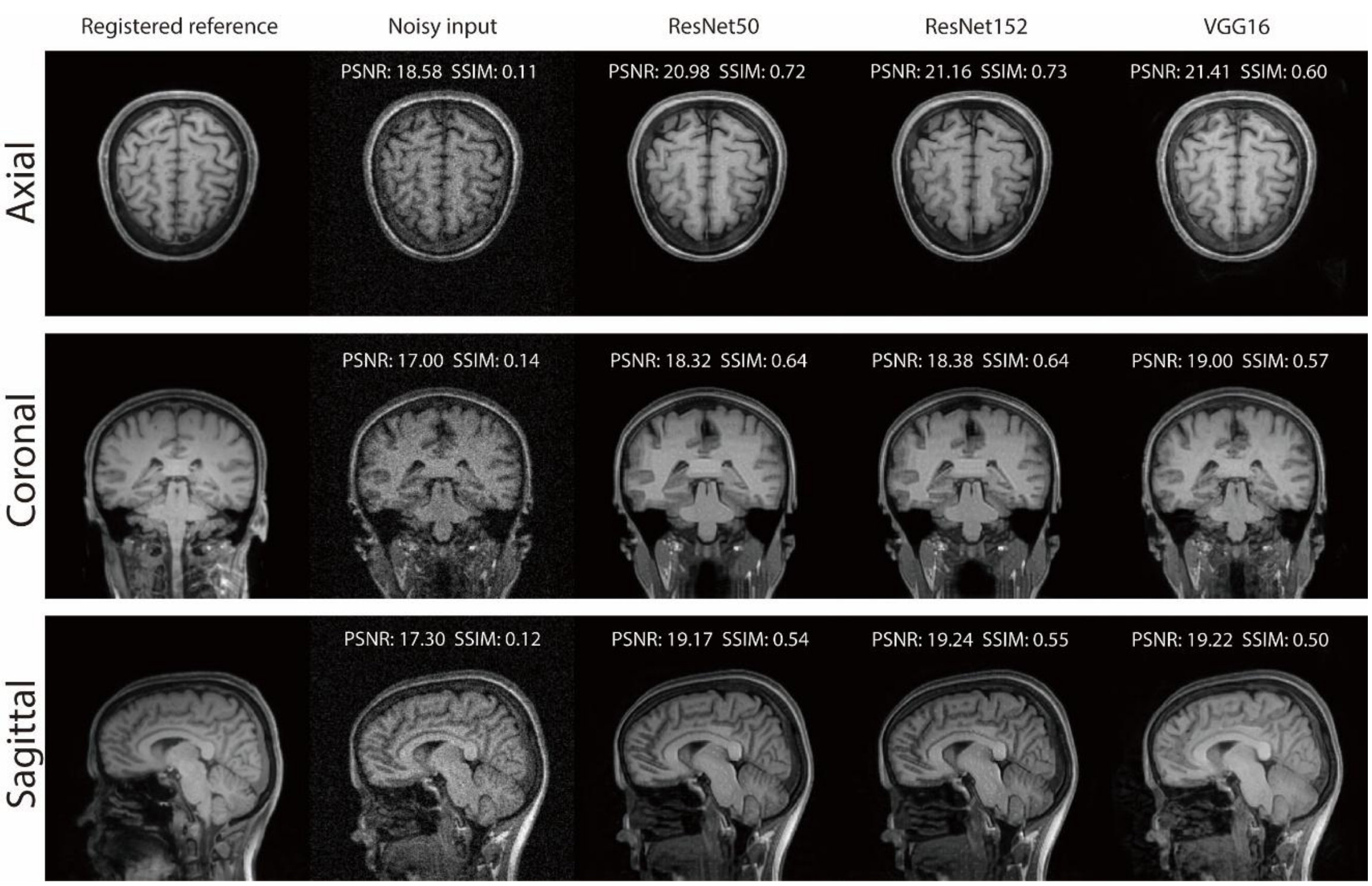

**Figure 2.** Reconstruction results of our RNST on the NACC dataset.

### 3.2 RNST reconstruction with unmatched images

The integration of an NST engine enables RNST to perform field-transfer reconstruction even in scenarios where only limited data is available or style images are not directly matched to the content images. In Table 3, we demonstrate comparison studies of RNST under matched guidance and frozen guidance setups. In the frozen guidance configuration, a fixed style image was used throughout the reconstruction, introducing greater content variation compared to the matched guidance setup.

Figures 3 and 4 illustrate reconstructed images from both matched and frozen guidance setups. Notably, RNST successfully maintained reconstruction quality despite the absence of an exact style-content match. The reconstructed images exhibited similar contrast and intensity to the high-field reference scans, with improved noise suppression. Figures 5 and S1 further analyze reconstructed images with zoomed-in views and error maps. These results validate RNST's effectiveness in field-transfer reconstructions without strict requirements for one-to-one style-content correspondence.

**Table 3.** Evaluation metrics of RNST reconstructions on the brain portion of the triple TSE scans.

| | | PSNR | | | SSIM | | |
|---|---|---|---|---|---|---|---|
| | | DA1 | DA2 | DA3 | DA1 | DA2 | DA3 |
| | | Brain portion, original scan | | | | | |
| | Input | 21.24 | 21.10 | 22.50 | 0.3464 | 0.3071 | 0.3215 |
| VGG16 | Match guide | 22.83 | 25.01 | 25.07 | 0.8178 | 0.8181 | 0.8210 |
| | Frozen guide | 22.68 | 24.60 | 24.96 | 0.7917 | 0.8041 | 0.8114 |
| ResNet50 | Match guide | 22.32 | 25.48 | 25.38 | 0.8219 | 0.8271 | 0.8228 |
| | Frozen guide | 22.00 | 25.14 | 24.87 | 0.8032 | 0.8265 | 0.8172 |
| ResNet152 | Match guide | 22.39 | 25.56 | 25.52 | 0.8296 | 0.8436 | 0.8308 |
| | Frozen guide | 21.99 | 25.10 | 24.89 | 0.7991 | 0.8220 | 0.8138 |
| | | Brain portion, additional noise | | | | | |
| | Input | 18.89 | 18.77 | 19.55 | 0.1288 | 0.1293 | 0.1005 |
| VGG16 | Match guide | 22.64 | 24.16 | 24.91 | 0.8183 | 0.8181 | 0.8212 |
| | Frozen guide | 22.73 | 23.81 | 25.38 | 0.8116 | 0.8114 | 0.8313 |
| ResNet50 | Match guide | 22.28 | 25.39 | 25.38 | 0.8109 | 0.8243 | 0.8210 |
| | Frozen guide | 21.85 | 24.95 | 24.84 | 0.7853 | 0.8198 | 0.8194 |
| ResNet152 | Match guide | 22.33 | 25.49 | 25.45 | 0.8203 | 0.8347 | 0.8268 |
| | Frozen guide | 21.81 | 24.94 | 24.89 | 0.7852 | 0.8105 | 0.8168 |

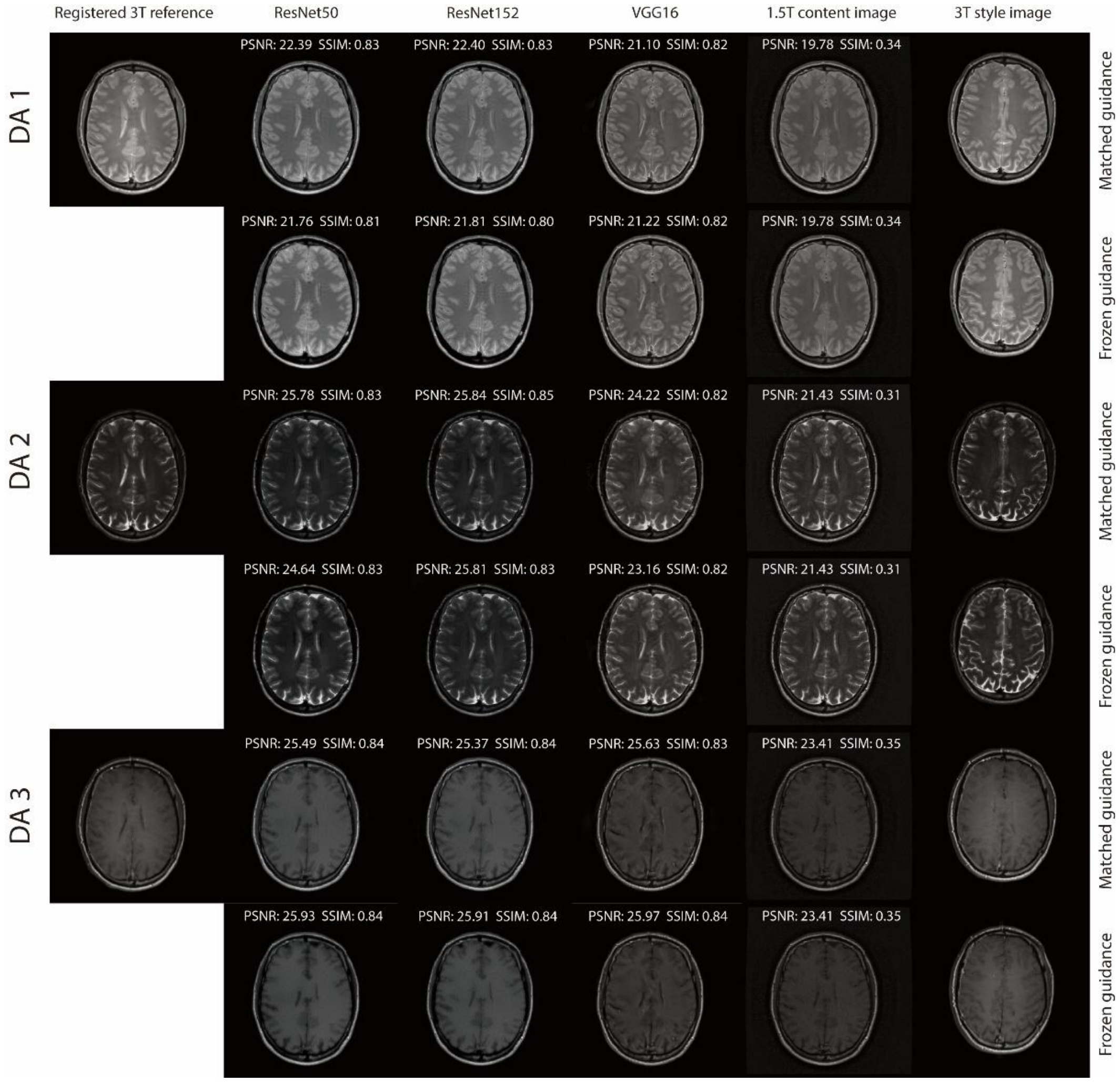

**Figure 3.** Reconstruction results of our RNST over the original scans on the triple TSE dataset.

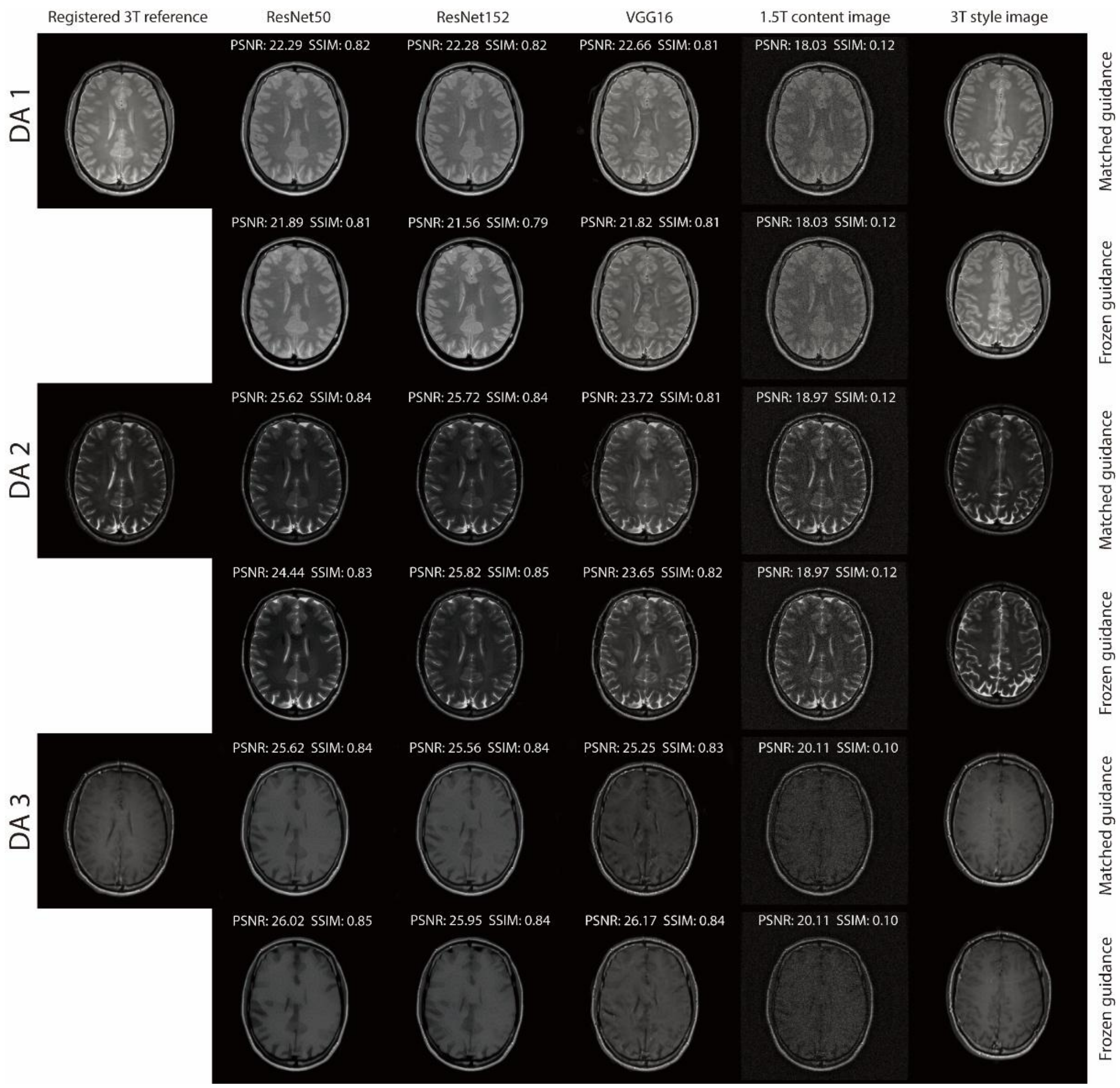

**Figure 4.** Reconstruction results of our RNST with additional noise on the triple TSE dataset.

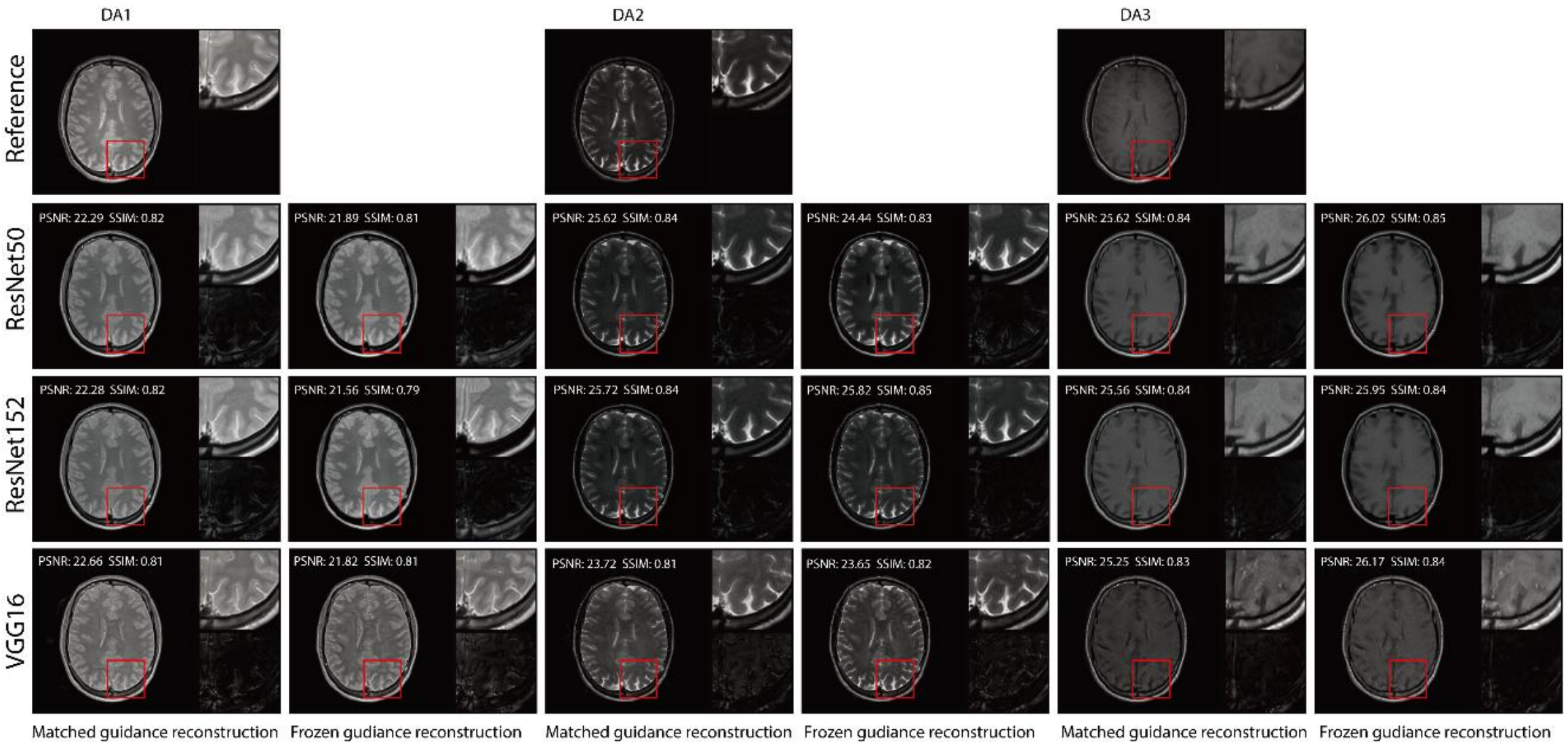

**Figure 5.** Visual illustration of RNST results for matched and frozen guidance setups for three DAs with additional noise on the triple TSE dataset. The image details highlighted in the red box in each figure were enlarged on the upper right side, with the corresponding error maps compared to the registered reference showing on the lower right side.

### 3.3 RNST performance along iterations

In Figures 6, S2, and S3, we demonstrate reconstruction samples along multiple iterations. Here, we present intermediate steps for iterations 1, 5 and 10, together with their intermediate evaluation metrics and error maps. The figures show that RNST performs reconstruction and noise reduction along iterations with better performance metrics as the iteration number increases. Overall, our experimental results highlight the capability of the RNST framework for limited data MRI reconstruction. This is especially helpful when scanning data is limited, precluding a large-scale deep neural network training with the potential to be further applied as an additional refinement to other reconstruction methods.

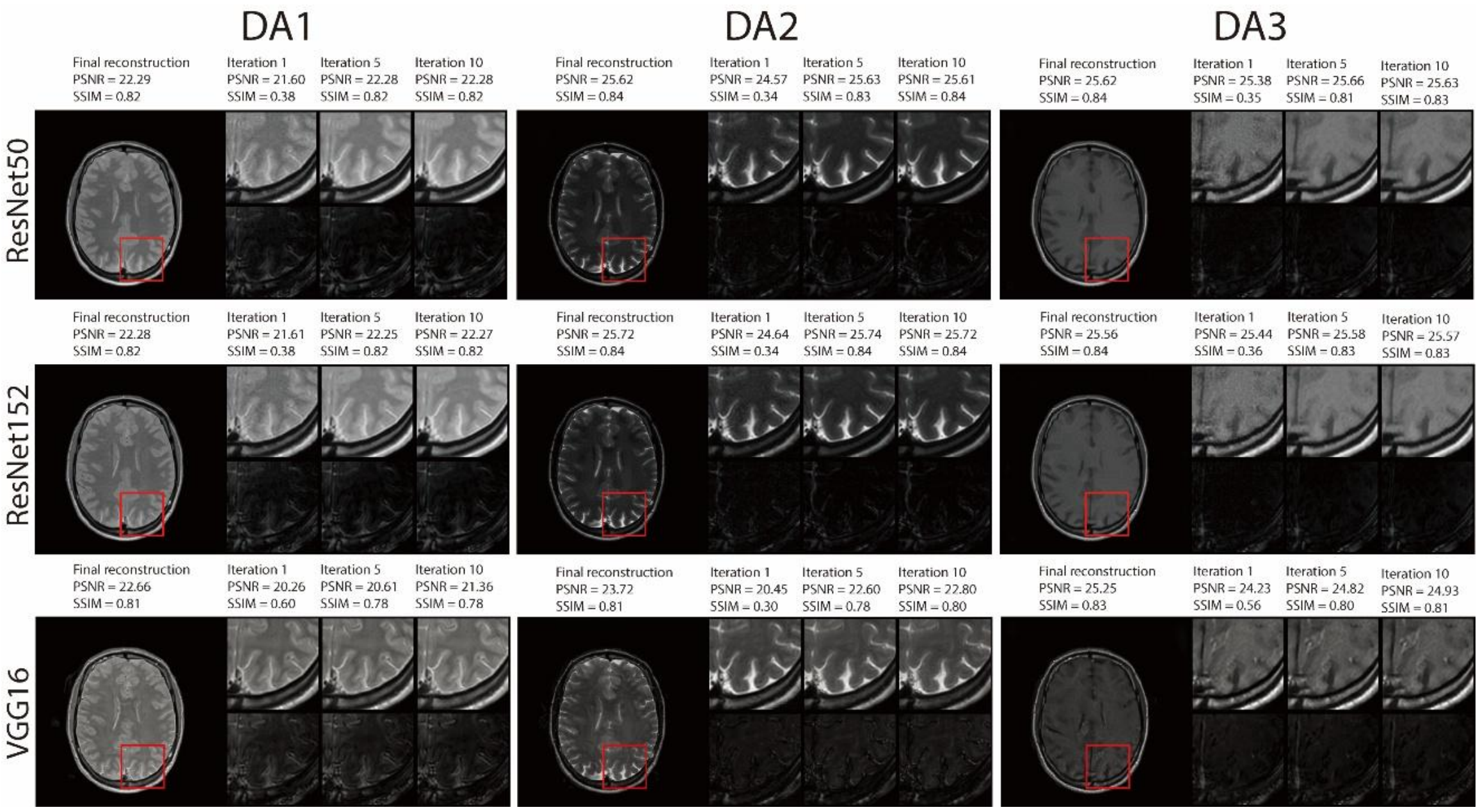

**Figure 6.** Visual illustration and quantitative metrics of the RNST reconstructions across iterations under matched guidance with additional noise on the triple TSE dataset. We present intermediate steps for iterations 1, 5 and 10, together with their intermediate evaluation metrics and error maps.

## 4 Conclusion

Deep learning-based MRI reconstruction frameworks typically require large-scale, task-specific datasets to achieve optimal performance. While these methods have achieved significant success, their applicability is often constrained in scenarios with limited data availability. The scarcity of adequately labeled medical data presents a major challenge, limiting the generalizability and practical deployment of these models. In this work, we propose a novel approach called Regularization by Neural Style Transfer (RNST) for MRI magnetic field-transfer reconstruction. RNST integrates a neural style transfer (NST) engine with a denoiser to generate high-field-quality images from noisy low-field inputs. By leveraging NST, RNST enables effective reconstruction with limited data, avoiding the need for extensive, task-specific training datasets.

Our results demonstrate that RNST consistently achieves superior image reconstruction quality across various experimental setups. Evaluation metrics presented in Tables 1 and 2 highlight the flexibility of RNST, illustrating that the NST engine can be applied across multiple deep neural network architectures while maintaining stable performance. Reconstructions performed in axial, coronal, and sagittal planes confirm the broad adaptability of RNST, eliminating the necessity for highly specific training datasets to ensure effective functionality. Additionally, experiments conducted on the TSE dataset reveal that RNST maintains robust reconstruction performance even in the presence of added noise corruption. Qualitative assessments in Figures 2-4 further substantiate these findings, showing that RNST successfully enhances image clarity, contrast, and intensity, producing reconstructions that closely resemble higher-field MRI references when compared to their lower-field counterparts.

A key advantage of RNST lies in its ability to perform field-transfer reconstruction with minimal data constraints while maintaining flexibility in style image selection. Comparison studies presented in Table 3 demonstrate that RNST achieves comparable reconstruction performance even when there is no exact style-content match. This underscores the model's robustness in scenarios where a direct one-to-one correspondence between style and content images is unavailable. Figures 5 and S1 provide further insight into this phenomenon, offering zoomed-in views that validate RNST's effectiveness in field-transfer reconstructions without rigid style-content pairing requirements. These findings indicate that RNST could be widely applied to diverse clinical settings where acquiring precisely matched reference images is impractical.

To gain further insights into the reconstruction performance across iterative steps, we conducted a detailed analysis of RNST's outputs at different iterations, as shown in Figures 6, S2, and S3. These evaluations reveal that RNST delivers promising results even within a few iterations. Notably, RNST with ResNet152 achieves a PSNR/SSIM of 25.74/0.84 (DA2) in Figure 6 after only five iterations. This demonstrates the efficiency of RNST's iterative process, where rapid improvements in reconstruction quality can be observed early in the optimization process. The ability to achieve high-fidelity reconstructions in a short number of iterations further enhances RNST's practical utility for real-time or near-real-time applications.

Despite these advancements, there remain several limitations and areas for future improvement. While RNST provides significant flexibility in terms of deep neural network architecture and eliminates the requirement for paired training datasets, it introduces additional challenges related to hyperparameter selection. Specifically, the choice of content and style extraction layers varies across different neural networks, necessitating careful tuning to optimize performance. To fully leverage the potential for advanced deep neural networks, the core problem remains in dataset quality and availability. The development of more advanced deep learning-based reconstruction techniques fundamentally relies on high-quality training data. Recent progress in conditional diffusion models [Zhan et al., 2024; Kim et al., 2023] has demonstrated significant potential in image synthesis, offering a promising path to address data scarcity. Additionally, techniques such as flow matching [Lipman et al., 2023; Lipman et al., 2024] suggest that diffusion-based models could serve as a viable solution for generating realistic medical images, as they are inherently rooted in probabilistic modeling and capable of capturing complex data distributions. Emerging research in general imaging models has begun incorporating synthetic data to further enhance performance [Shin et al., 2018; Kirillov et al., 2023], providing a compelling direction for future work. Utilizing such generative models could further improve reconstruction accuracy and robustness, particularly in settings where real-world data acquisition is constrained. Furthermore, the effectiveness of RNST in clinical practice will ultimately depend on its ability to preserve diagnostically relevant details while avoiding hallucination artifacts. Future studies should focus on evaluating RNST's performance on clinical datasets containing subtle abnormalities, such as small strokes or metastatic lesions, to further assess its diagnostic reliability.

In conclusion, this work introduces Regularization by Neural Style Transfer (RNST) as an innovative solution for MRI magnetic field-transfer reconstruction. RNST demonstrates superior performance across various imaging configurations, showcasing its flexibility in integrating an NST engine and its robustness in scenarios without exact style-content alignment. By addressing the challenge of field-transfer reconstruction with limited data, RNST represents a promising framework that could significantly impact the field of MRI reconstruction, offering a scalable and effective approach for improving image quality in resource-limited environments.

## Conflict of Interest

*The authors declare that the research was conducted in the absence of any commercial or financial relationships that could be construed as a potential conflict of interest.*

## Author Contributions

GS: Conceptualization, Methodology, Software, Formal analysis, Writing – original draft, Writing – review & editing. YZ: Methodology, Software, Writing – original draft. ML: Methodology, Software, Formal analysis, Writing – original draft. RM: Methodology, Software, Writing – original draft. HJ: Formal analysis, Writing – review & editing. SBA: Writing – review & editing. CWF: Formal analysis, Writing – review & editing. SA: Conceptualization, Methodology, Writing – review & editing. XZ: Conceptualization, Methodology, Formal analysis, Writing – original draft, Writing – review & editing, Project administration, Funding acquisition.

## Funding

The author(s) declare that financial support was received for the research, authorship, and/or publication of this article. The present study has received assistance from the Rajen Kilachand Fund for Integrated Life Science and Engineering.

## Acknowledgments

This work was supported by the Rajen Kilachand Fund for Integrated Life Science and Engineering. We would like to thank the Boston University Photonics Center for technical support.

The NACC database is funded by NIA/NIH Grant U24 AG072122. NACC data are contributed by the NIA-funded ADRCs: P30 AG062429 (PI James Brewer, MD, PhD), P30 AG066468 (PI Oscar Lopez, MD), P30 AG062421 (PI Bradley Hyman, MD, PhD), P30 AG066509 (PI Thomas Grabowski, MD), P30 AG066514 (PI Mary Sano, PhD), P30 AG066530 (PI Helena Chui, MD), P30 AG066507 (PI Marilyn Albert, PhD), P30 AG066444 (PI David Holtzman, MD), P30 AG066518 (PI Lisa Silbert, MD, MCR), P30 AG066512 (PI Thomas Wisniewski, MD), P30 AG066462 (PI Scott Small, MD), P30 AG072979 (PI David Wolk, MD), P30 AG072972 (PI Charles DeCarli, MD), P30 AG072976 (PI Andrew Saykin, PsyD), P30 AG072975 (PI Julie A. Schneider, MD, MS), P30 AG072978 (PI Ann McKee, MD), P30 AG072977 (PI Robert Vassar, PhD), P30 AG066519 (PI Frank LaFerla, PhD), P30 AG062677 (PI Ronald Petersen, MD, PhD), P30 AG079280 (PI Jessica Langbaum, PhD), P30 AG062422 (PI Gil Rabinovici, MD), P30 AG066511 (PI Allan Levey, MD, PhD), P30 AG072946 (PI Linda Van Eldik, PhD), P30 AG062715 (PI Sanjay Asthana, MD, FRCP), P30 AG072973 (PI Russell Swerdlow, MD), P30 AG066506 (PI Glenn Smith, PhD, ABPP), P30 AG066508 (PI Stephen Strittmatter, MD, PhD), P30 AG066515 (PI Victor Henderson, MD, MS), P30 AG072947 (PI Suzanne Craft, PhD), P30 AG072931 (PI Henry Paulson, MD, PhD), P30 AG066546 (PI Sudha Seshadri, MD), P30 AG086401 (PI Erik Roberson, MD, PhD), P30 AG086404 (PI Gary Rosenberg, MD), P20 AG068082 (PI Angela Jefferson, PhD), P30 AG072958 (PI Heather Whitson, MD), P30 AG072959 (PI James Leverenz, MD).

## Supplementary Material

Please see the Supplementary Material section.

## Data Availability Statement

The NACC dataset is available at: https://naccdata.org/requesting-data/data-request-process. The TSE scanning data that supports the findings are available from the Boston Medical Center. Restrictions apply to the availability of these data, which were used under license for this study. The codes of this study are openly available at: https://github.com/GuoyaoShen/RNST-Regularization_by_NST.

## Supplementary Material

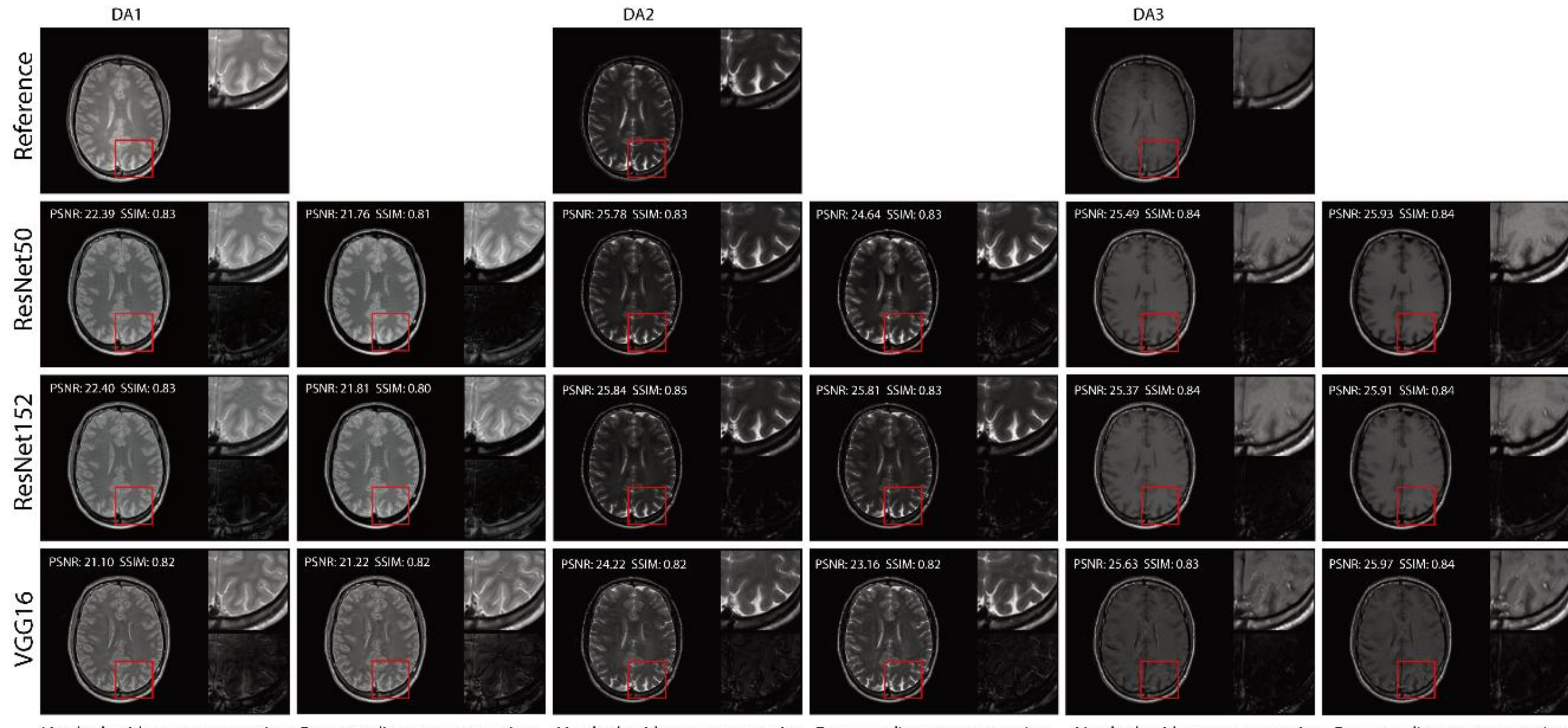

**Supplementary Figure S1.** Visual illustration of RNST results for matched and frozen guidance setups for three DAs on the original scans on the triple TSE dataset. The image details highlighted in the red box in each figure were enlarged on the upper right side, with the corresponding error maps compared to the registered reference showing on the lower right side.

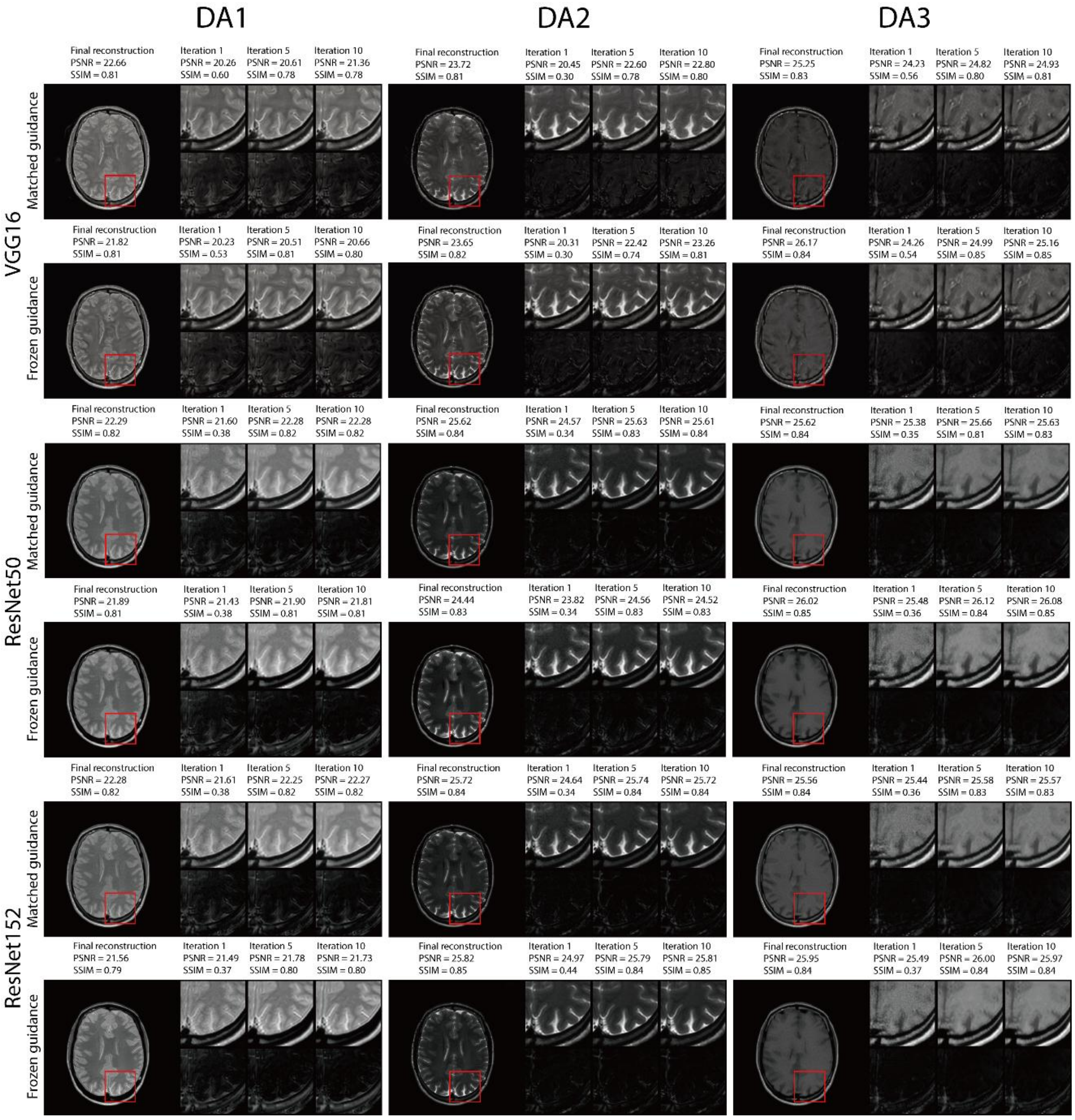

**Supplementary Figure S2.** Visual illustration and quantitative metrics of the RNST reconstructions across iterations with additional noise on the triple TSE dataset. We present intermediate steps for iterations 1, 5 and 10, together with their intermediate evaluation metrics and error maps.

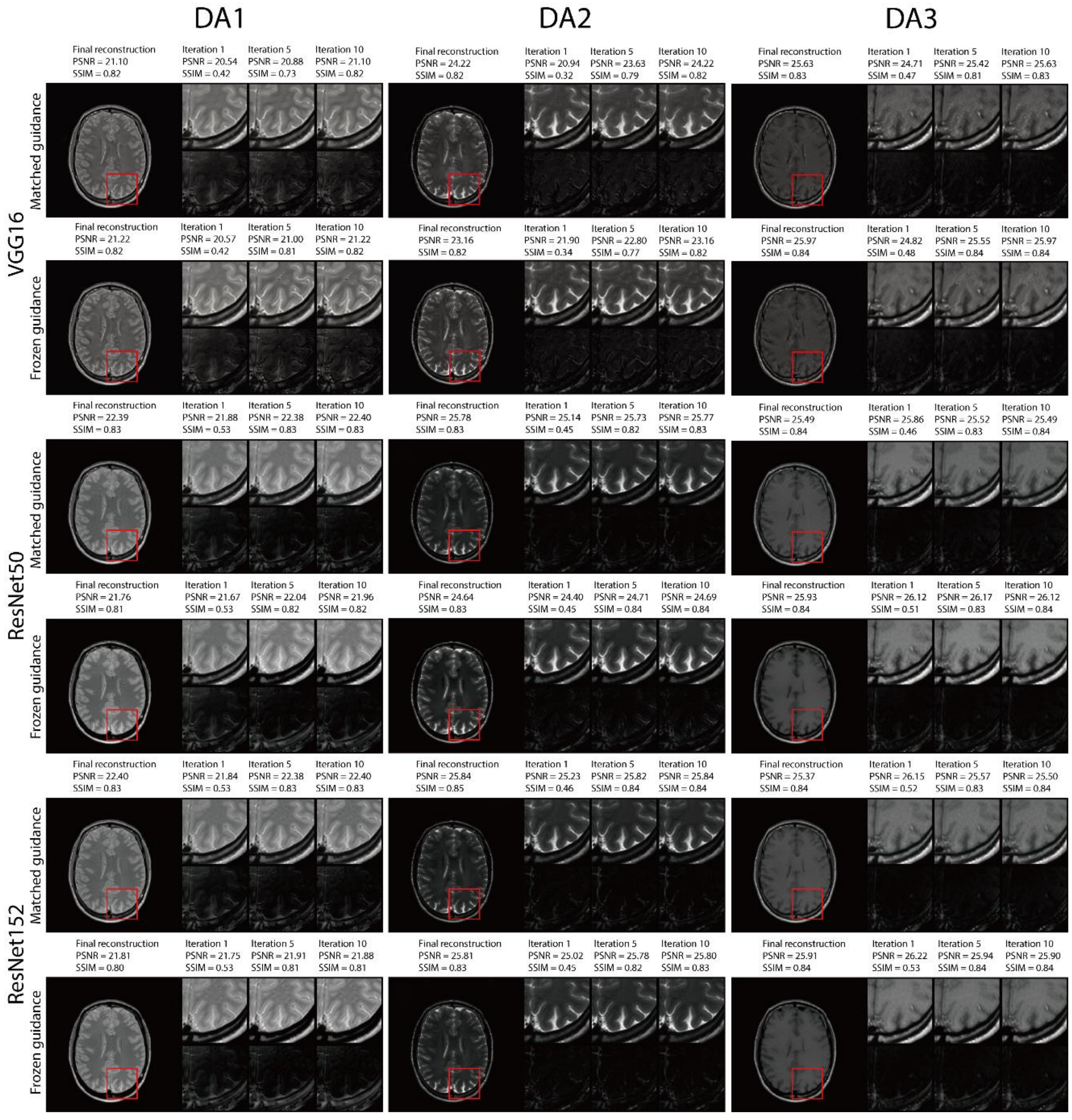

**Supplementary Figure S3.** Visual illustration and quantitative metrics of the RNST reconstructions across iterations on the original scans on the triple TSE dataset. We present intermediate steps for iterations 1, 5 and 10, together with their intermediate evaluation metrics and error maps.